\documentclass[11pt,a4paper]{article}
\usepackage{authblk}
\usepackage[hyperref]{eacl2021}
\usepackage{times}
\usepackage{latexsym}
\usepackage{graphicx}

\usepackage{amsmath}
\usepackage{mathtools}
\usepackage{multicol}
\usepackage{multirow}

\usepackage{microtype}

\aclfinalcopy 
\def\aclpaperid{134} 

\title{CD$^{2}$CR: Co-reference Resolution Across Documents and Domains}

\author[1,3,4]{\textbf{James Ravenscroft}}
\author[5]{\textbf{Arie Cattan}}
\author[2]{\textbf{Amanda Clare}}
\author[5]{\textbf{Ido Dagan}}
\author[1,3]{\textbf{Maria Liakata}}

\affil[1]{Centre for Scientific Computing, University of Warwick, CV4 7AL, United Kingdom}
\affil[2]{Department of Computer Science, Aberystwyth University, SY23 3DB, United Kingdom}
\affil[3]{Alan Turing Institute, 96 Euston Rd, London, NW1 2DB, United Kingdom}
\affil[4]{Filament AI, 1 King William St, London, EC4N 7BJ, United Kingdom}
\affil[5]{Computer Science Department, Bar-Ilan University, Ramat-Gan,
Israel}

\date{}

\begin{document}
\maketitle
\begin{abstract}

Cross-document co-reference resolution (CDCR) is the task of identifying and linking mentions to entities and concepts across many text documents. Current state-of-the-art models for this task assume that all documents are of the same type (e.g. news articles) or fall under the same theme. However, it is also desirable to perform CDCR across different domains (type or theme). A particular use case we focus on in this paper is the resolution of entities mentioned across scientific work and newspaper articles that discuss them. Identifying the same entities and corresponding concepts in both scientific articles and news can help scientists understand how their work is represented in mainstream media. We propose a new task and English language dataset for cross-document cross-domain co-reference resolution (CD$^2$CR). The task aims to identify links between entities across heterogeneous document types. We show that in this cross-domain, cross-document setting, existing CDCR models do not perform well and we provide a baseline model that outperforms current state-of-the-art CDCR models on CD$^2$CR. Our data set, annotation tool and guidelines as well as our model for cross-document cross-domain co-reference are all supplied as open access open source resources.
\end{abstract}

\section{Introduction}
Cross-document co-reference resolution (CDCR) is the task of recognising when multiple documents mention and refer to the same real-world entity or concept. CDCR is a useful NLP process that has many downstream applications. For example, CDCR carried out on separate news articles that refer to the same politician can facilitate inter-document sentence alignment required for stance detection and natural language inference models. Furthermore, CDCR can improve information retrieval and multi-document summarisation by grouping documents based on the entities that are mentioned within them.

Recent CDCR work \cite{Dutta2015,Barhom2019,Cattan2020} has primarily focused on resolution of entity mentions across news articles. Despite differences in tone and political alignment, most news articles are relatively similar in terms of grammatical and lexical structure.  Work based on modern transformer networks such as BERT \cite{Devlin2018} and ElMo \cite{Peters:2018} have been pre-trained on large news corpora and are therefore well suited to news-based CDCR \cite{Barhom2019}.

However, there are  cases where CDCR across documents from different domains (i.e. that differ much more significantly in style, vocabulary and structure) is useful. One such example is the task of resolving references to concepts across scientific papers and related news articles. This can help scientists understand how their work is being presented to the public by mainstream media or facilitate fact checking of journalists' work~\cite{Wadden2020}.
A chatbot or recommender that is able to resolve references to current affairs in both news articles and user input could be more effective at suggesting topics that interest the user. Finally, it may be helpful for e-commerce companies to know when product reviews gathered from third party websites refer to one of their own listings. The work we present here focuses on the first cross-document, cross-domain co-reference-resolution (CD$^2$CR) use case, namely co-reference resolution between news articles and scientific papers.

The objective of CD$^2$CR is to identify co-referring entities from documents belonging to different domains. 
In this case co-reference resolution is made more challenging by the differences in language use (lexical but also syntactic) across the different domains. 
Specifically, authors of scientific papers aim to communicate novel scientific work in an accurate and unambiguous way by using precise scientific terminology. Whilst scientific journalists also aim to accurately communicate novel scientific work, their work is primarily funded by newspaper sales and thus they also aim to captivate as large an audience as possible. Therefore journalists tend to use simplified vocabulary and structure, creative and unexpected writing style, slang, simile, metaphor and exaggeration to make their work accessible, informative and entertaining in order to maximise readership \cite{Louis2013}. 

Success at the CD$^2$CR task in this setting is dependent on context sensitive understanding of how the accessible but imprecise writing of journalists maps on to precise terminology used in scientific writing. For example, a recent study has found that ``convalescent plasma derived from donors who have recovered from COVID-19 can be used to treat patients sick with the disease'' \footnote{DOI: 10.1101/2020.03.16.20036145}. A news article\footnote{\url{https://tinyurl.com/ycnq9xg7}} discussing this work says that ``...blood from recovered Covid-19 patients in the hope that transfusions...[can help to treat severely ill patients]" . In this example the task is to link `blood' to `convalescent plasma' and `recovered Covid-19 patients' to `donors'. These cross-document, cross-domain co-reference chains can be used as contextual anchors for downstream analysis of the two document settings via tasks such as natural language inference, stance detection and frame analysis. 

\noindent The contributions in this paper are the following:

\begin{itemize}
    \item A novel task setting for CDCR that is more challenging than those that already exist due to linguistic variation between different domains and document types  (we call this CD$^2$CR).
    \item An open source English language CD$^2$CR dataset with 7602 co-reference pair annotations over 528 documents and detailed 11 page annotation guidelines (section \ref{sec:data_collection}).
    \item A novel annotation tool to support ongoing data collection and annotation for CD$^2$CR including a novel sampling mechanism for calculating inter-annotator agreement (Section \ref{sec:annotation_ui}). 
    \item A series of experiments on our dataset using different baseline models and an in-depth capability-based evaluation of the best-performing baseline (Section \ref{sec:results})
\end{itemize}

\section{Related Work}

\subsection{Co-reference Resolution}
\label{sec:related_coreference}
Intra-document co-reference resolution is a well understood task with mature training data sets \cite{weischedel2013ontonotes} and academic tasks \cite{Recasens2010SemEval2010T1}. The current state of the art model by \citet{Joshi2020} is based on \citet{lee-etal-2017-end,lee2018} and uses a modern BERT-based \cite{Devlin2018} architecture. Comparatively, CDCR, which involves co-reference resolution across multiple documents, has received less attention in recent years~\cite{Bagga1998,Rao2010,Dutta2015,Barhom2019}. \citet{Cattan2020} jointly learns both entity and event co-reference tasks, achieving current state of the art performance for CDCR,  and as such provides a strong baseline for experiments in CD$^2$CR. 
Both \citet{Cattan2020} and \citet{Barhom2019} models are trained and evaluated using the ECB+ corpus \cite{Cybulska2014} which contains news articles annotated with both entity and event mentions.


\subsection{Entity Linking}
Entity Linking (EL) focuses on alignment of mentions in documents to resources in an external knowledge resource~\cite{Ji2010} such as SNOMED CT\footnote{\url{https://tinyurl.com/yy7g4ttz}} or DBPedia\footnote{\url{https://wiki.dbpedia.org/}}. EL is challenging due to the large number of pairwise comparisons between document mentions and knowledge resource entities that may need to be carried out. \citet{Raiman2018} provide state of the art performance by building on \citet{ling2015}'s work in which an entity type system is used to limit the number of required pairwise comparisons to related types. \citet{Yin2019} achieved comparable results using a graph-traversal method to similarly constrain the problem space to candidates within a similar graph neighbourhood. EL can be considered a narrow sub-task of CDCR since it cannot resolve novel and rare entities or pronouns~\cite{Shen2015}. Moreover EL's dependency on expensive-to-maintain external knowledge graphs is also problematic when limited human expertise is available. Given these limitations, EL is inappropriate within our task setting, hence our CDCR-based approach.

\subsection{Semantic Specialisation} 
\label{sec:related_semantic}

Like earlier static vector language models, contextual language models such as BERT \cite{Devlin2018} and ElMo \cite{Peters:2018} use distributional knowledge~\cite{Harris1954} inherent in large text corpora to learn context-aware word embeddings that can be used for downstream NLP tasks. However, these models do not learn about formal lexical constraints, often conflating different types of semantic relatedness~\cite{ponti2018, lauscher2019}. This is a weakness of all distributional language models that is particularly problematic in the context of CD$^2$CR for entity mentions that are related but not co-referent (e.g. ``Mars" and ``Jupiter") as shown in section \ref{sec:results}.

A number of solutions have been proposed for adding lexical knowledge to static word embeddings~\cite{yu2014, Wieting2015, ponti2018} but contextual language models have received comparatively less attention. Lauscher et al \shortcite{lauscher2019} propose adding a lexical relation classification step to BERT's language model pre-training phase to allow the model to integrate both lexical and distributional knowledge. Their model, LIBERT, has been shown to facilitate statistically-significant performance boosts on a variety of downstream NLP tasks. 
\vspace{-0.2cm}
\section{Dataset creation}
Our dataset is composed of pairs of news articles and scientific papers gathered automatically (Section \ref{sec:data_collection}). Our annotation process begins by obtaining summaries of the news and science document pairs (extractive news summaries and scientific abstracts, respectively) (Section \ref{sec:summarisation}). Candidate co-reference pairs from each summary-abstract pair are identified and scored automatically. 
(Section \ref{sec:task_gen}).
Candidate co-reference pairs are then presented to human annotators via a bespoke annotation interface for scoring (Section \ref{sec:annotation_ui}). Annotation quality is measured on an ongoing basis as new candidates are added to the system (Section \ref{sec:iaa}).

\subsection{Data Collection}
\label{sec:data_collection}
We have developed a novel data set that allows us to train and evaluate a CD$^2$CR model. The corpus is approximately 50\% the size of the ECB+ corpus (918 documents) \cite{Cybulska2014}  and is split into training, development and test sets (statistics for each subset are provided in Table \ref{tab:corpus_stats}). Each pair of documents consists of a scientific paper and a newspaper article that discusses the scientific work. In order to detect pairs of documents, we follow the approach of \cite{Ravenscroft2018}, using approximate matching of author name and affiliation metadata, date of publishing and exact DOI matching where available to connect news articles to scientific publications.

\begin{table}
\begin{tabular}{|l|lll|}
\hline
\textbf{Subset }& \textbf{Documents }& \textbf{Mentions }& \textbf{Clusters } \\
Train & 300       & 4,604     & 426   \\
Dev    & 142       & 1,821     & 199   \\
Test   & 86        & 1,177     & 101       \\\hline
\end{tabular}
\caption{Total individual documents, mentions, co-reference clusters of each subset excluding singletons.}
\label{tab:corpus_stats}
\end{table}

We built a web scraper that scans for new articles from the `Science' and `Technology' sections of 3 well-known online news outlets (BBC, The Guardian, New York Times) and press releases from Eurekalert, a widely popular scientific press release aggregator. Once a newspaper article and related scientific paper are detected, the full text from the news article and the scientific paper abstract and metadata are stored. Where available the full scientific paper content is also collected. We ran the scraper between April and June 2020 collecting news articles and scientific papers including preprints discussing a range of topics such as astronomy, computer science and biology (incl. coverage of COVID-19). New relevant content is downloaded and ingested into our annotation tool (see Section \ref{sec:annotation_ui}) on an ongoing basis as it becomes available.

\subsection{Article Summarisation}
\label{sec:summarisation}
Newspaper articles and scientific papers are long and often complex documents, usually spanning multiple pages, particularly the latter. Moreover the two document types differ significantly in length. Comparing documents of such uneven length is a difficult task for human annotators.  We also assume that asking human annotators to read the documents in their entirety to identify co-references would be particularly hard with a very low chance for good inter-annotator agreement (IAA). 
We therefore decided to simplify the task by asking annotators to compare summaries of the newspaper article (5-10 sentences long) and the scientific paper (abstract). 



For each document pair, we ask the annotators to identify co-referent mentions between the scientific paper abstract  and a summary of the news article that is of similar length (e.g. 5-10 sentences). Scientific paper abstracts act as a natural summary of a scientific work and have been used as a strong baseline or even a gold-standard in scientific summarisation tasks \cite{Liakata2013}. Furthermore, abstracts are almost always available rather than behind paywalls like full text articles. For news summarisation, we used a state-of-the-art extractive model \cite{Grenander2019} to extract sentences forming a summary of the original text. This model provides a summary de-biasing mechanism preventing it from focusing on specific parts of the full article, preserving the summary's informational authenticity as much as possible.

The difference in style between the two documents is preserved by both types of summary since abstracts are written in the same scientific style as full papers and the extractive summaries use verbatim excerpts of the original news articles.




\subsection{Generation of pairs for annotation}
\label{sec:task_gen}

\begin{figure*}
    \centering
    \includegraphics[width=\textwidth]{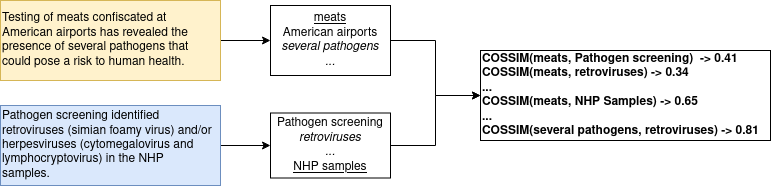}
    \caption{Illustration of the generation process for pairs of potentially co-referring expressions, left boxes represent related news summary (top) and abstract (bottom), co-referent entity pairs in middle boxes shown with same formatting (underline,italic).}
    \label{fig:cossim_process}
\end{figure*}

To populate our annotation tool, we generate pairs of candidate cross-document mentions to be evaluated by the user. Candidate mentions are identified by using spaCy \cite{Honnibal2017} for the recognition of noun phrases and named entities from each input document pair (abstract-news summary). For each pair of documents, pairs of all possible mention combinations are generated and stored for annotation.

In any given pair of documents, the majority of mention pairs ($M_0$, $M_1$) generated automatically in this way will not co-refer thus resulting in a vastly imbalanced dataset and also running the risk of demotivating annotators. 
To ensure that annotators are exposed to both positive and negative examples, we use a similarity score to rank examples based on how likely they are to co-refer. The first step in generating a similarity score $s$ is to concatenate each abstract-news-summary pair together: ``summary [SEP] abstract" into a pre-trained $\text{BERT}_{\text{large}}$ model. Then we take the mean of the word vectors that correspond to the mention spans within the documents and calculate the cosine similarity of these vectors. We find that this BERT-based similarity score performs well in practice. We also use it in combination with a thresholding policy as one of our baseline models in Section \ref{sec:models}. 

\subsection{Annotation Tool \& Interface}
\label{sec:annotation_ui}

We developed an open source annotation tool\footnote{https://github.com/ravenscroftj/cdcrtool} that allows humans to identify cross-document co-reference between each pair of related documents. Whilst designing this tool, we made a number of decisions to simplify the task and provide clear instructions for the human annotators in order to encourage consistent annotation behaviour.

\begin{figure*}
    \centering
    \includegraphics[width=0.8\textwidth]{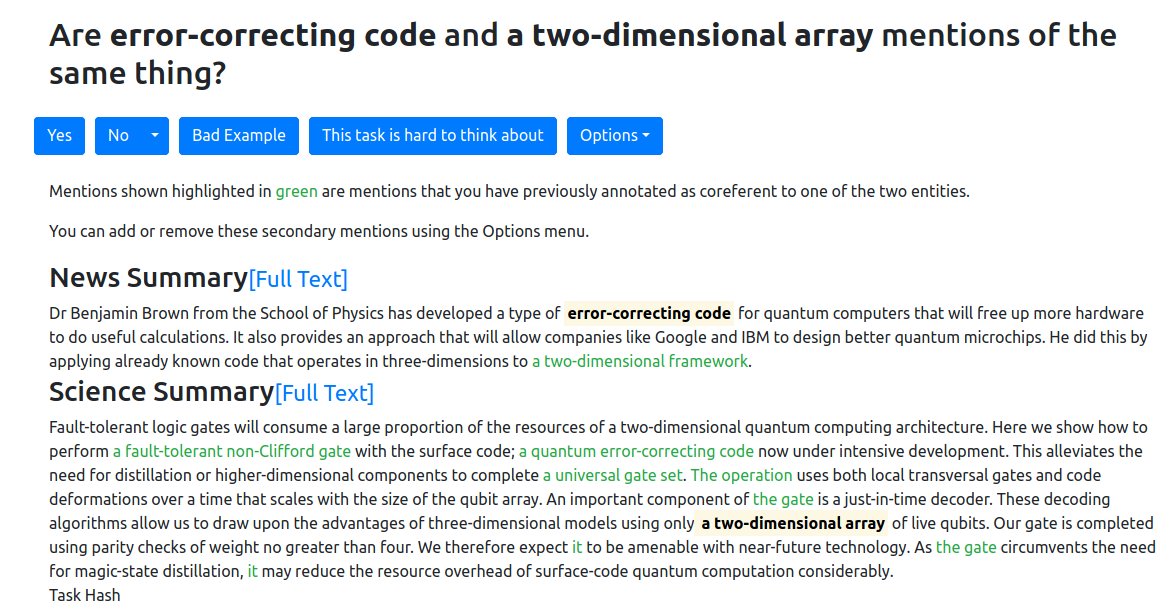}
    \caption{An example of a cross-document co-reference task presented within our annotation tool.}
    \label{fig:cdcrtool}
    \vspace{-0.4cm}
\end{figure*}

To maximise the quality and consistency of annotations in our corpus, we simplified the task as much as possible for the end user. Annotation tasks were framed as a single yes or no question: ``Are \textbf{x} and \textbf{y} mentions of the same entity?".  
Mentions in context were shown in bold font whereas mentions already flagged as co-referent were shown in green. This enabled annotators to understand the implications for existing co-reference chains before responding (see Figure \ref{fig:cdcrtool}).
Questions were generated and ranked via our task generation pipeline (see Section \ref{sec:task_gen} above).

We added two additional features to our annotation interface to improve annotators' experience and to speed up the annotation process. Firstly, if the candidate pair is marked as co-referent, the user is allowed to add more mentions to the coreference cluster at once. Secondly, inspired by \cite{Li2020}, if the automatically shown mention pair is not co-referent, the user can select a different mention that is co-referent. 

The upstream automated mention detection mechanism can sometimes introduce incomplete or erroneous mentions, leading to comparisons that don't make sense or that are particularly difficult. Therefore, annotators can also move or resize the mention spans they are annotating. 

We use string offsets of mention span pairs to tokens to check that they do not overlap with each other in order to prevent the creation of duplicates. 
Figure \ref{fig:cossim_process} shows an illustrated example of the generation pipeline for mention pairs.

\subsection{Annotation Protocol}
\label{sec:iaa}
We recruited three university-educated human annotators and 
provided them with detailed annotation guidelines for the resolution of yes/no questions on potentially co-referring entities in pairs from the ordered queue described above. By default each entity pair resolution is carried out once, allowing us to quickly expand our data set. However, we pseudo-randomly sample 5\% of mention pairs in order to calculate inter-annotator-agreement (IAA) and make sure that data collected from the tool is consistent and suitable for modelling. New entity pairs for IAA are continually sampled as new document pairs and mention tuples are added to the corpus by the web scraper (Section \ref{sec:data_collection}). The annotation system puts mention pairs flagged for IAA first in the annotation queue. Thus, all annotators are required to complete IAA comparisons before moving on to novel mention pairs. This allows us to ensure that all annotators are well represented in the IAA exercise. To avoid annotators being faced with a huge backlog of IAA comparisons before being able to proceed with novel annotations, we also limited the number of comparisons for IAA required by each user to a maximum of 150 per week.

\subsection{Task Difficulty and Annotator Agreement}

We anticipated that annotation of the CD$^2$CR corpus would be difficult in nature due to its dependencies on context and lexical style. We invited users to provide feedback regularly to help us refine and clarify our guidelines and annotation tool in an iterative fashion. Users could alert us to examples they found challenging by flagging them as difficult in the tool. Qualitative analysis of the subset of `difficult' cases showed that the resolution of mention pairs is often perceived by annotators as difficult when:

\begin{itemize}
    \item Deep subject-matter-expertise is required to understand the mentions, e.g. is ``jasmonic acid" the same as ``regulator cis ‐(+)‐12‐oxophytodienoic acid".
    \item Mentions involve non-commutable set membership ambiguity e.g. ``Diplodocidae" and ``the dinosaurs"
    \item Mentions are context dependent e.g. ``the struggling insect" and ``the monarch butterfly".
\end{itemize}

This feedback prompted the introduction of highlighting for existing co-reference chains in the user interface (as described in section \ref{sec:annotation_ui} above) to make it easier to tell when non-commutable set membership would likely introduce inconsistencies into the dataset. For mention pairs requiring subject-matter-expertise, annotators were encouraged to research the terms online. For context sensitive mention pairs, annotators were encouraged to read the full news article and full scientific paper in order to make a decision. 

In our 11 page annotation guidelines document (appendix) we describe the use of our annotation tool and illustrate some challenging CD$^2$CR tasks and resolution strategies. For example precise entities mentioned in the scientific document may be referenced using ambiguous exophoric mentions in the news article (e.g. `a mountain breed of sheep' vs `eight ovis aries').  Our guidelines require resolving these cases based on the journalist's intent (e.g. `a mountain breed' refers to the `ovis aries' sheep involved in the experiment).

We evaluated the final pairwise agreement between annotators using Cohen's Kappa \cite{Cohen1960} ($\kappa_{\text{cohen}}$) and an aggregate `n-way' agreement score using Fleiss' Kappa \cite{Fleiss1971} ($\kappa_{\text{fleiss}}$). Pairwise $\kappa_{\text{cohen}}$ is shown in table \ref{table:kappa_scores} along with the total number of tasks each annotator completed. Annotator 3 (A3) shows the most consistent agreement with the other two annotators. Our Fleiss' Kappa analysis of tasks common across the three annotators gave $\kappa_{\text{fleiss}}=0.554$. We note that Fleiss' Kappa is a relatively harsh metric and values, like ours, between 0.41 and 0.60 are considered to demonstrate 'moderate agreement'\cite{Landis1977}. We also carried out Fleiss' Kappa analysis on the subset of mention pairs that were completed by all annotators and were also marked as difficult by at least one user (180 mention pairs in total). We found that for this subset of pairs, $\kappa_{\text{fleiss}}=0.399$ which is considered to be fair agreement\cite{Landis1977}.

\begin{table}
\begin{tabular}{|l|llll|}
\hline
   & \# Annotations & A1    & A2         & A3    \\ \hline
A1 & 10,685          & -     & 0.492      & 0.600 \\
A2 & 3,051           & 0.492 & \textbf{-} & 0.500 \\
A3 & 9,847           & 0.600 & 0.500      & -     \\ \hline
\end{tabular}
\caption{Number of Annotations and Pairwise Cohen's Kappa scores $\kappa_{\text{cohen}}$ between annotators.}
\label{table:kappa_scores}
\vspace{-0.4cm}
\end{table}


\section{Model}
\label{sec:models}

Below we describe several baseline models including state of the art CDCR models that we used to evaluate how well current approaches can be used in our CD$^2$CR task setting.

\subsection{BERT Cosine Similarity (BCOS) Baseline}
\label{sec:bert}
In this model we calculate the cosine-similarity between embeddings of the two mentions in context ($M_0, M_1$) encoded using a pre-trained BERT model as discussed above in section \ref{sec:task_gen}. We define a thresholding function $f$ to decide if $M_0$ and $M_1$ are co-referent ($f(x)=1$) or not ($f(x)=0$):
\[
    f(x)= 
\begin{dcases}
    1,& \text{if } \text{COSSIM}(M_0,M_1)\geq t\\
    0,              & \text{otherwise}
\end{dcases}
\]

During inference, we pass this function over all pairs $M_0,M_1$ and infer missing links such that if $f(A,B)=1$ and $f(B,C)=1$ then $f(A,C)=1$.

\begin{figure}[!ht]
    \centering
    \includegraphics[width=0.4\textwidth]{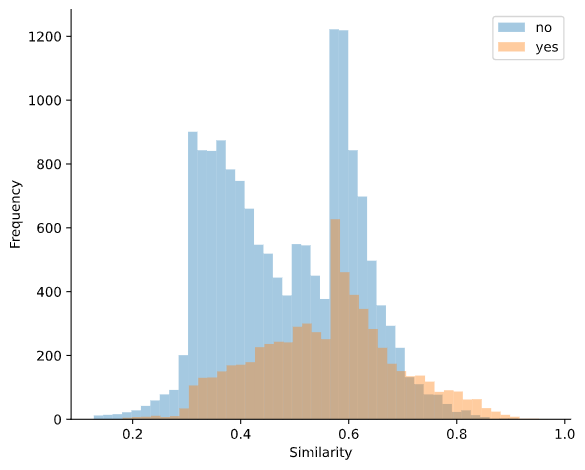}
    \caption{BERT Cosine Similarity frequency distribution for co-referent (Yes) and non-co-referent (No) mention pairs in the CD$^2$CR corpus. }
    \label{fig:sim_distributions}
    \vspace{- 0.3 cm}
\end{figure}

Based on Figure \ref{fig:sim_distributions}, we test values in increments of 0.01 between 0.3 and 0.8 inclusive for threshold cut off $t$. We evaluated the baseline by measuring its accuracy at predicting co-reference in each mention pair in the $CD^2CR$ development set. The best performance was attained when $t=0.65$.  A visualisation of the BERT Cosine Similarity distributions of co-referent and non co-referent annotated mention pairs can be seen in Figure \ref{fig:sim_distributions}. 

Co-referent mention pairs tend to have a slightly higher BERT cosine similarity than non co-referent mention pairs but there is significant overlap of the two distributions suggesting that in many cases BERT similarity is too simplistic a measure.

\subsection{Entities Only Baseline (CA)}
\label{sec:cattan_baseline}

We use a state-of-the-art model \cite{Cattan2020} (CA) for cross-document co-reference resolution. In this model, each document is separately encoded using a RoBERTa encoder (without fine-tuning) to get contextualized representations for each token. Then, similarly to the within-document co-reference model by \citet{lee-etal-2017-end}, the mention spans are represented by the concatenation of four vectors: the vectors of the first and last token in the span, an attention-weighted sum of the span token vectors, and a feature vector to encode the span width. Two mention representations are then concatenated and fed to a feed-forward network to learn a likelihood score for whether two mentions co-refer. At inference time, agglomerative clustering is used on the pairwise scores to form coreference clusters.


The CA model is trained to perform both event and entity recognition on the ECB+ corpus \cite{Cybulska2014} 
In our setting there is no event detection subtask so, for fair comparison, we pre-train the CA model on ECB+ entity annotations only and evaluate it on our new CD$^2$CR task to see how well it generalises to our task setting.

\subsection{CA + Fine-Tuned (CA-FT) Baseline}
Here we aim to evaluate whether fine tuning the CA model from section \ref{sec:cattan_baseline} using the CD$^2$CR corpus can improve its performance in the new task setting. The CA model is first trained on the ECB+ corpus in the manner described above. We then further fine-tune the feed-forward model (without affecting the RoBERTa encoder) on the CD$^2$CR corpus for 10 epochs with early stopping. Pseudo-random sub-sampling is carried out on the training set to ensure a balance of co-referent and non-co-referent mention pairs.

\subsection{CA - Vanilla (CA-V) Baseline}
\label{sec:ca-vr-model}
Here we aim to evaluate whether training the CA model on the CD$^2$CR dataset from the RoBERTa baseline without first training on the ECB+ corpus allows it to fit well to the new task setting.
We re-initialise the CA encoder (Section \ref{sec:cattan_baseline}) using weights from $\text{RoBERTa}$ \cite{Liu2019} and randomly initialise the remaining model parameters. We then train the model on the CD$^2$CR corpus for up to 20 epochs with early stopping with pseudo-random sub-sampling as above.

\subsection{CA - SciBERT (CA-S) Baseline }

This model is the same as CA-V but we replace the $\text{RoBERTa}$ encoder with SciBERT \cite{Beltagy2019}, a version of BERT pre-trained on scientific literature in order to test whether the scientific terms and context captured by SciBERT improve performance at the CD$^2$CR task compared to $\text{RoBERTa}$. Similarly to CA-V in section \ref{sec:ca-vr-model}, we initialise the BERT model with weights from $\text{SciBERT}_\text{scivocab-uncased}$ \cite{Beltagy2019} and randomly initialise the remaining model parameters, training on the  CD$^2$CR corpus for up to 20 epochs with early stopping.

\section{Results and Discussion}

\begin{table}[b]
\begin{tabular}{|p{1.18cm}|p{0.5cm}lp{0.5cm}|p{0.5cm}p{0.5cm}p{0.5cm}|}
\hline
\multirow{2}{*}{\textbf{Model}} & \multicolumn{3}{c|}{\textbf{MUC}}     & \multicolumn{3}{c|}{\textbf{$\text{B}^3$}} \\ \cline{2-7} 
                                & \textbf{P} & \textbf{R} & \textbf{F1} & \textbf{P}      & \textbf{R}     & \textbf{F1}     \\ \hline
\textbf{BCOS}          & 0.42       &\textbf{0.94}       & 0.58        & 0.01            & 0.45           & 0.00            \\
\textbf{CA}                     & 0.41       & 0.51       & 0.46        & \textbf{0.39}            & 0.33           & 0.35            \\
\textbf{CA-V}                  & 0.50       & 0.69       & \textbf{0.58}      & 0.35            & 0.57           & \textbf{0.44}           \\
\textbf{CA-FT}                  & 0.47       & 0.71       & 0.52        & 0.30            & \textbf{0.62}           & 0.41            \\
\textbf{CA-S}                 & \textbf{0.58}       & 0.46       & 0.51        & 0.32            & 0.53           & 0.39            \\ \hline
\end{tabular}

\caption{MUC and $B^3$ results from running baseline models on CD$^2$CR test subset, BCOS threshold=0.65 }
\label{tab:results}
\end{table}


\label{sec:results}

We evaluate each of the model baselines described in section \ref{sec:models} above on the test subset of our CD$^2$CR corpus. Results are shown in table \ref{tab:results}.

For the purposes of evaluation, we use named entity spans from the manually annotated CD$^2$CR as the ``gold standard" in all experiments rather than using the end-to-end Named Entity Recognition capabilities provided by some of the models. We evaluate the models using the metrics described by \citet{Vilain1995} (henceforth MUC) and  \citet{Bagga1998} (henceforth $B^3$). MUC F1, precision and recall are defined in terms of pairwise co-reference relationships between each mention. $B^3$, F1, precision and recall are defined in terms of presence or absence of specific entities in the cluster. When measuring $B^3$, we remove entities with no co-references (singletons) from the evaluation to avoid inflation of results \cite{Cattan2020}.

\begin{table*}
\begin{tabular}{|p{1.75cm}|p{1.45cm}|p{2cm}|p{10cm}|}
\hline
Test Type                         & Co-referent? & Pass Rate \& Total Tests & Example test case and outcome for test case                                                                                                                                                                    \\ \hline
\multirow{2}{1.75cm}{Anaphora and Exophora resolution} & Yes         & 47.1\% (16/34) & \begin{tabular}[c]{@{}l@{}}M1: ...to boost \textbf{the struggling insect}'s numbers...\textit{
[PASS]}\\ M2: the annual migration of the \textbf{monarch butterfly}... \end{tabular}             \\ 
    \cline{2-4}
                                                  & No          & 76.5\% (26/34) & \begin{tabular}[c]{@{}l@{}}M1: ...\textbf{monarchs} raised in captivity...\textit{
[FAIL]}\\ M2: ...rearing wild-caught \textbf{monarchs} in an indoor environment...\end{tabular}       \\ \hline
\multirow{2}{1.75cm}{Subset relationship resolution}   & Yes         & 24.3\%  (9/37)  & \begin{tabular}[c]{@{}l@{}}M1: ...it was in fact a hive of \textbf{human activity}...\textit{
[FAIL]}\\ M2: ...this region for \textbf{Pre-Columbian cultural developments}...\end{tabular} \\ \cline{2-4}
                                                  & No          & 60.0\% (18/30) & \begin{tabular}[c]{@{}l@{}}M1: ... \textbf{the carnivore's skull}...\textit{
[FAIL]}\\ M2: ... \textbf{the gigantic extinct Agriotherium africanum}\end{tabular}                 \\ \hline 
\multirow{2}{1.75cm}{Para-phrase resolution}          & Yes         & 33.3\% (13/39) & \begin{tabular}[c]{@{}l@{}}M1: ...\textbf{a giant short-faced bear}...\textit{
[PASS]}\\ M2: ...\textbf{the gigantic extinct Agriotherium africanum}...\end{tabular}                                                                \\ \cline{2-4}
                                                  & No          & 80.5\% (29/36)  & \begin{tabular}[c]{@{}l@{}}M1: ...half the energy that \textbf{existing techniques} require\textit{
[FAIL]}\\ M2: ...the lack of \textbf{efficient catalysts} for ammonia synthesis\end{tabular}                        \\ \hline

\end{tabular}

\caption{A breakdown of specific tests carried out on CA-V model against three challenging types of relationships found in the $CD^2CR$ corpus. [PASS] or [FAIL] indicates CA-V model correctness. Pass Rate is mathematically equivalent to Recall for test sets.}
\label{tab:unit_tests}
\vspace{-0.3cm}
\end{table*}


The threshold baseline (BCOS) gives the highest MUC recall but also poor MUC precision and poorest $B^3$ precision. The $B^3$ metric is highly specific with respect to false-positive entity mentions and strongly penalises BCOS for linking all non-coreferent pairs with $COSSIM(M_0,M_1) \geq 0.65 $. Furthermore, Fig. \ref{fig:sim_distributions} shows that a thresholding strategy is clearly sub-optimal given that there is a significant overlap of co-referent and non-co-referent pairs with only a small minority of pairs at the top and bottom of the distribution that do not overlap. Therefore, despite its promising MUC F1 score, it is clear that BCOS is not useful in practical terms.

\begin{figure}
    \centering
    \includegraphics[width=0.4\textwidth]{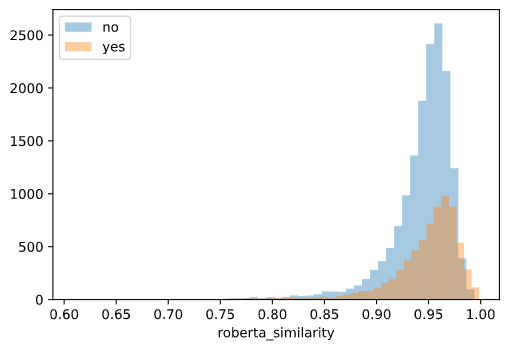}
    \caption{RoBERTa Cosine Similarity frequency distribution for co-referent (Yes) and non-co-referent (No) mention pairs in the CD$^2$CR corpus. Distribution is compressed between ~0.8 and ~1.0.}
    \label{fig:roberta_distributions}
\end{figure}

Whilst our thresholding baseline above uses BERT, RoBERTa is used by \citet{Cattan2020} as the basis for their state-of-the-art model and thus for our models based on their work. Although the two models have the same architecture, RoBERTa has been shown to outperform BERT at a range of tasks~\cite{Liu2019}.
However, as shown in Figure \ref{fig:roberta_distributions}, the cosine similarity distribution of mention pair embeddings produced by RoBERTa is compressed to use a smaller area of the potential distribution space compared to that of BERT (Figure \ref{fig:sim_distributions}). This compression of similarities may imply a reduction in RoBERTa's ability to discriminate in our task setting. \citet{Liu2019} explain that their byte-pair-encoding (BPE) mechanism, which expands RoBERTa's sub-word vocabulary and simplifies pre-processing, can reduce model performance for some tasks, although this is not further explored in their work. We leave further exploration of RoBERTa's BPE scheme and its effects on the CD$^2$CR task setting to future work.


All of the models specifically trained on the CD$^2$CR corpus (CA-V, CA-FT, CA-S) outperform the CA model by a large margin. Furthermore, the CA-V model (without pre-training on ECB+ corpus) outperforms the CA-FT model (with ECB+ pre-training) by 6\% MUC and 3\% B$^3$. These results suggest that the CD$^2$CR task setting is distinct from the CDCR and ECB+ task setting and that this distinction is not solvable with fine-tuning.  

In terms of both MUC and B$^3$, CA-S performs much worse than CA-V suggesting that SciBERT embeddings are less effective than $\text{RoBERTa}$ embeddings in this task setting.  We hypothesise that SciBERT's specialisation towards scientific embeddings may come at the cost of significantly worse news summary embeddings when compared to those produced by $\text{RoBERTa}$.

We next evaluate our best performing CD$^2$CR baseline model (CA-V) at the entity resolution CDCR task using the ECB+ test corpus, to see how well it generalises to the original CDCR task. Results are presented in \ref{table:ecb_results} along-side Cattan et al's original model results (CA). The CA-V model still shows good performance, despite a small drop, when compared to the original CA model. The drop in $B^3$ F1 is more pronounced than MUC but is still broadly in line with other contemporary CDCR systems \cite{Cattan2020}.  The CA-V model demonstrates a promising ability to generalise beyond our corpus to other tasks and reveals an interesting correspondence between CDCR and CD$^2$CR settings.

\begin{table}[hbt]
\begin{tabular}{|p{1.18cm}|p{0.5cm}lp{0.5cm}|p{0.5cm}p{0.5cm}p{0.5cm}|}
\hline
\multirow{2}{*}{\textbf{Model}} & \multicolumn{3}{l|}{\textbf{MUC}}     & \multicolumn{3}{l|}{\textbf{$B^3$}} \\ \cline{2-7} 
                                & \textbf{P} & \textbf{R} & \textbf{F1} & \textbf{P}      & \textbf{R}     & \textbf{F1}     \\ \hline
\textbf{CA}                     & 0.86      & 0.82      & 0.84       & 0.63           & 0.68          & 0.65           \\
\textbf{CA-V}                   & 0.82      & 0.81      & 0.81       & 0.56           & 0.53          & 0.55   \\      \hline
\end{tabular}
\caption{MUC and $B^3$ results from running the CD$^2$CR baseline model (CA-V) on ECB+ dataset compared with original \citet{Cattan2020} (CA).}
\label{table:ecb_results}
\end{table}

\begin{figure}
    \centering
    \includegraphics[width=0.4\textwidth]{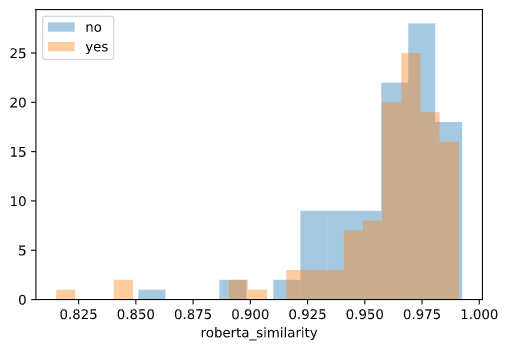}
    \caption{RoBERTa-based mention pair similarity frequency distribution for test examples from Table \ref{tab:unit_tests}. 'yes' and 'no' for 'co-referent' and 'not-co-referent' respectively }
    \label{fig:test_distributions}
\end{figure}

Finally, the best model (CA-V) is analysed using a series of challenging test cases inspired by Ribeiro et al \shortcite{Ribeiro2020}. These test cases were created using 210 manually annotated mention-pairs found in the test subset of the $CD^2CR$ corpus according to the type of relationship illustrated (Anaphora \& Exophora, Subset relationships, paraphrases). We collected a balanced set of 30-40 examples of both co-referent and non-coreferent-but-challenging pairs for each type of relationship (exact numbers in Table \ref{tab:unit_tests}). We then recorded whether the model correctly predicted co-reference for these pairs. The results along with illustrative examples of each relationship type are shown in Table \ref{tab:unit_tests}. The results suggest that the model is better at identifying non-co-referent pairs than co-referent pairs and that it struggles with positive co-referent mentions for all three types of relationship. The model struggles to relate general reader-friendly descriptions of entities from news articles to precise and clinical descriptions found in scientific papers. The model often successfully identifies related concepts such as `the carnivore's skull' and `Agriotherium africanum'. However it is unable to deal with the complexity of these relationships and appears to conflate `related' with `co-referent', which is likely due to lack of lexical knowledge as we discussed in section \ref{sec:related_semantic}. Figure \ref{fig:test_distributions} shows significant overlap between co-referent and non-co-referent RoBERTa-based cosine similarities, which can also be observed for the wider corpus in Figure \ref{fig:roberta_distributions}, but is especially bad for these test examples. This overlap suggests that disentangling these pairs is likely to be a challenging task for the downstream classification layer in the CA-V model. These challenges are less likely to occur in homogeneous corpora like ECB+ where descriptions and relationships remain consistent in detail and complexity.

\section{Conclusion}

We have defined cross-document, cross-domain co-reference resolution (CD$^2$CR), a special and challenging case of cross-document co-reference resolution for comparing mentions across documents of different types and/or themes.  We have constructed a specialised CD$^2$CR annotated dataset, available, along with our annotation guidelines and tool, as a free and open resource for future research. We have shown that state-of-the-art CDCR models do not perform well on the CD$^2$CR dataset without specific training. Furthermore, even with task-specific training, models perform modestly and leave room for further research and improvement. Finally, we show that the understanding of semantic relatedness offered by current generation transformer-based language models may not be precise enough to reliably resolve complex linguistic relationships such as those found in CD$^2$CR as well as other types of co-reference resolution and relationship extraction tasks. The use of semantic enrichment techniques (such as those discussed in Section \ref{sec:related_semantic}) to improve model performance in the CD$^2$CR task should be investigated as future work.


\section*{Acknowledgments}
This work was supported by The Alan Turing Institute under the EPSRC grant EP/N510129/1 and the University of Warwick's CDT in Urban Science under EPSRC grant EP/L016400/1.


\bibliography{eacl2021}
\bibliographystyle{acl_natbib}

\end{document}